\documentclass[sn-mathphys]{sn-jnl}

 \normalbaroutside
 
\usepackage{times}
\usepackage{array}
\usepackage{pgf}
\usepackage{tikz}
\usepackage{pgfplots}
\usepackage{pgfplotstable}
\usepgflibrary{shapes.geometric}
\usepgfplotslibrary{groupplots}

\pgfplotsset{compat=1.7}

\usepackage{pdfpages}

\begin{document}

\includepdf{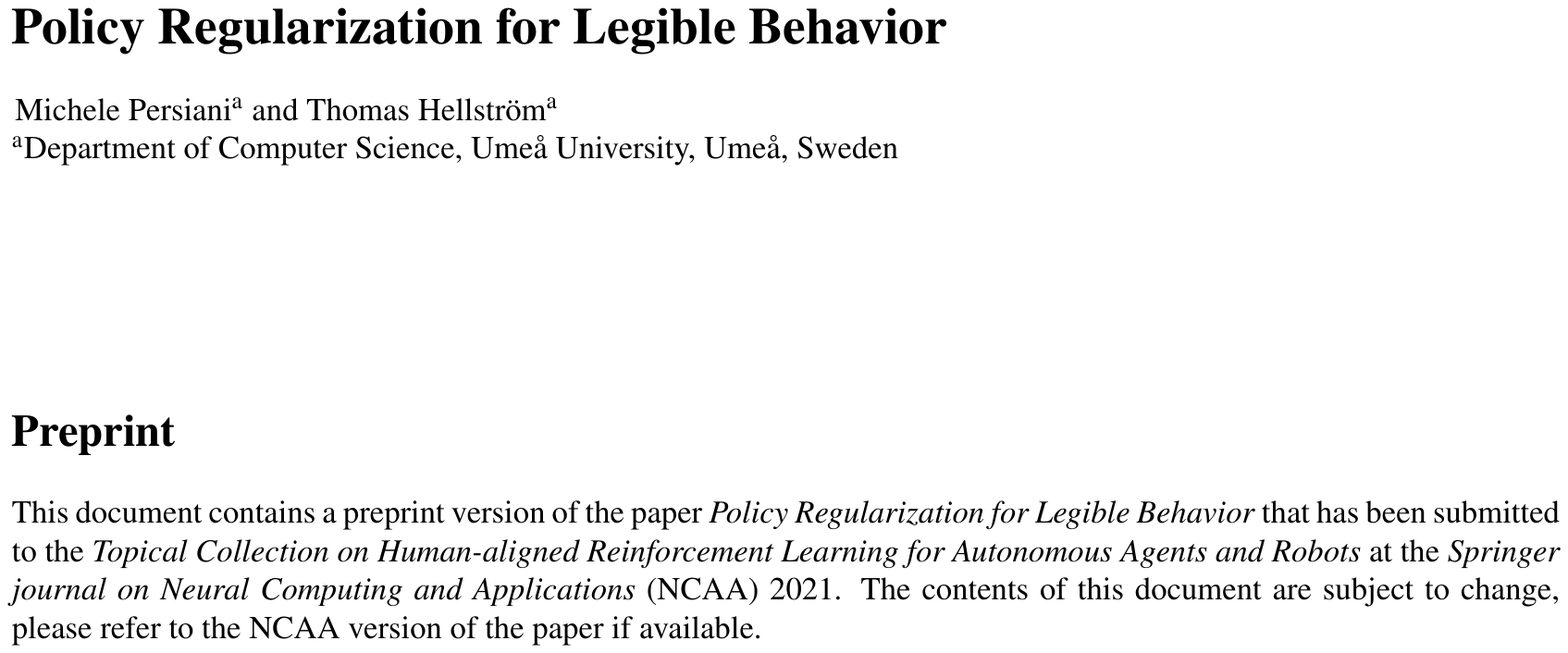}

\title[Policy Regularization for Legible Behavior]{Policy Regularization for Legible Behavior}

\author*[1]{\fnm{Michele} \sur{Persiani}}\email{michelep@cs.umu.se}

\author[1]{\fnm{Thomas} \sur{Hellstr\"om}}\email{thomas.hellstrom@umu.se}

\affil*[1]{\orgdiv{Department of Computing Science}, \orgname{Ume{\aa} University}, \orgaddress{\city{Ume{\aa}}, \country{Sweden}}}

\abstract{
In Reinforcement Learning interpretability generally means to provide insight into the agent's mechanisms such that its decisions are understandable by an expert upon inspection. This definition, with the resulting methods from the literature, may however fall short for online settings where the fluency of interactions prohibits deep inspections of the decision-making algorithm. To support interpretability in online settings it is useful to borrow from the Explainable Planning literature methods that focus on the legibility of the agent, by making its intention easily discernable in an observer model. As we propose in this paper, injecting legible behavior inside an agent's policy doesn't require  modify components of its learning algorithm. Rather, the agent's optimal policy can be regularized for legibility by evaluating how the policy may produce observations that would make an observer infer an incorrect policy. In our formulation, the decision boundary introduced by legibility impacts the states in which the agent's policy returns an action that has high likelihood also in other policies. In these cases, a trade-off between such action, and legible/sub-optimal action is made.
}


\maketitle

\section{Introduction}

As widely agreed in Explainable Artificial Intelligence, well-functioning collaboration between humans and artificial agents requires transparency~\cite{Sule19}. Agents should not only perform their assigned tasks  efficiently and accurately, but should also make sure that the humans in their operative context understand their intentions and actions. 

Facilitating intention recognition through a behavior that is understandable by a human observer has several advantages \cite{hellstrom2018understandable}. For example, in human-robot interaction signaling the robot's intention increases collaborators' trust in the robot, safety, and fluency of interactions because aiding collaborators to predict what the robot is doing or will do \cite{schaefer2017communicating,chang2018effects,sciutti2018humanizing}, and in conditions of shared control allows to mediate, arbitrate, and guide the interaction \cite{losey2018review} by informing the user about the robot's intended action. In applications for autonomous vehicles simple solutions augmenting the driver understanding of the car's intentional state, like sharing its goal, is sufficient to increase trustworthiness and acceptability of the autonomous driving system, as well as acceptance of higher levels of automation \cite{verberne2012trust}. In addition, recent developments in technologies for virtual or mixed reality are further enabling and enhancing methods for intentionality in physical robots, by allowing to plot and manipulate the robots' intentional states in the virtual 3D world \cite{walker2018communicating}.

\begin{figure}[t]
    \centering
    \includegraphics[width=.8\linewidth]{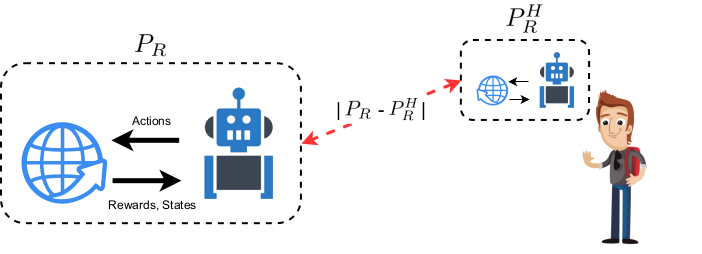}
    \caption{$P_R$: an RL agent interacting with its environment. $P_R^H$: model of the expectations that an observer has about the agent. The goal of interpretable behavior is to keep the distance $\mid P_R - P_R^H\mid$ low, signifying that the agent's behavior effectively matches the observer's expectations.}
    \label{fig:front}
\end{figure}

Given the importance of intentions during interactions with artificial agents, it is therefore becoming relevant to combine methods that allow to express intentions with techniques generating highly performing behavior. The online creation of behavior of which intention is easily discernable or that is furnished with congruent explanations is addressed in Explainable Planning under the umbrella of \textit{interpretable behavior}, where several methods to regularize behavior for explicability \cite{kulkarni2019explicable}, predictability \cite{zhang2017plan} or legibility \cite{dragan2013legibility,persiani2021probabilistic} have been proposed. These techniques relate to an implicit communication of intention by making it transparent to its user, and is in contrast with explanations that instead is an explicit communication. Transparency is achieved by interacting with a user observer model. For example, legibility skews plan trajectories such that their goal is easily discernable, explicability makes sure that observations have at least one associated complete plan, or predictability reduces the amount of possible future possible trajectories. 

While a substantial amount of formalizations of interpretable behavior exists in the Explainable Planning literature, there is very little related work for the framework of Reinforcement Learning (RL). RL has been shown to produce powerful agents for a variety of domains (including robotics, games or recommender systems) often surpassing human performance, however, the RL framework still lacks formalization about creating interpretable agents as intended in Explainable Planning, and mostly borrows its definition of interpretability from the Machine Learning (ML) literature. This definition is more concerned into making the decision taken by the algorithm explainable by a domain expert upon inspection in an offline setting, rather than to enable interpretability online during collaborations, therefore resulting unsuitable in fulfilling the needs of transparency of online interactions.

There is therefore still a large untapped potential in adapting methods for interpretability to RL. This would also provide valuable input for research in explainability that at the moment contemplates advanced methods such as those based on neural networks mostly as black boxes generating behavior that is optimal yet highly uninterpretable from a human perspective \cite{puiutta2020explainable}. To this purpose, in this paper we translate the legibility criteria from Explanable Planning to the RL framework as a measure of discernability of policy, that we loosely equal to the agent's intention. As we propose, injecting legibility inside an agent's policy doesn't require to modify components of the learning algorithm. We rather suggest to evaluate how the optimal policy may produce state-action pairs that would make the observer infer a wrong policy, to later find a trade-off that minimizes those while remaining consistent to the original policy.

\section{Background}

Since RL borrows the term ``interpretability'' mostly from the ML literature \cite{alharin2020reinforcement,du2019techniques}, merging the terminology from Explainable Planning and Reinforcement Learning could create some confusion. In ML interpretability generally means to provide insight into the agent's mechanisms such that its decisions are understandable by an expert upon inspection \cite{du2019techniques}. This can be achieved firstly by translating the classifiers' latent features responsible for its decisions into a space that is interpretable, and then compute explanations on that space \cite{roscher2020explainable}. In RL, \cite{mott2019towards} for example proposes to use attention to visualize which features the deep Q-network attends when taking decisions, while \cite{liu2018toward} trains linear tree models on Deep Q-networks to obtain corresponding interpretable models. See \cite{alharin2020reinforcement} for a survey of this type of techniques applied to RL. 

These techniques for interpretability have been shown useful in many ML application domains  by giving insight into models' decisions. They have, for example, been successful in  health-care \cite{stiglic2020interpretability}, and societal (eg. decisions regarding loans, hiring, risks, etc.) applications. However, they may be less suitable in domains characterized by real-time interaction, such as in human-robot interaction, where the fluency of the interaction prohibits deep inspections of the decision making algorithm. Also, while the produced explanations in terms of relevant features could be understood by an expert, they may be unsuitable for users who are uninformed of the underlying models, and more focused on common sense reasoning. People are in general very good at forming hypotheses on intentions and beliefs explaining an observed behavior through what is referred to a theory of mind reasoning \cite{rutherford2004effect}. However, it has been commonly shown how the behavior of advanced agents operating at human level, such as in competitive games, are often beyond human intuition and highly inexplicable \cite{perez2020adopting,firestone2020performance}. Especially for such cases, but also in general, it is therefore necessary to regularize artificial agents towards behaviors compatible with common sense reasoning, while maintaining their high performance.

To this end, in this paper we refer to interpretability as intended in planning, where an agent behavior is interpretable when an observer can easily discern what the agent is doing by understanding its intention \cite{chakraborti2019explicability}. Also when applied to RL, this definition conforms better to real-time interaction in the presence of an observer that could be either passive or part of a larger collaborating agent, such as a human. As previously introduced, in this context a multitude of definitions capturing smaller aspects of interpretability have been used. Each aspect expresses different types of expectations that an eventual observer has on the agent, such as expectations about its goal \cite{dragan2013legibility}, expectations about entire future trajectories \cite{zhang2017plan}, or expectations towards a communication model \cite{macnally2018action}. While there is a lot of variety in the models and theories leveraged by all this techniques, it can be generally shown that this set of methods requires an expectation model that is a second-order theory of mind focused on the observer's inferences about the agent \cite{hellstrom2018understandable,chakraborti2019explicability}, and that interpretable behavior can be seen as minimizing the distance between the estimated model possessed by the observer and the true model of the agent (see Figure~\ref{fig:front}). The agent's behavior is interpretable whenever conforming with the expectations casted by the second-order model, and uninterpretable when not conforming \cite{hellstrom2018understandable}.

In agents applications the second-order theory of mind is the model that the agent thinks the observer is using to interpret its behavior and can have many forms, for example, in \cite{zhang2017plan} it is a label predicting whether a human observer is understanding the agent, while in \cite{chakraborti2017plan} is a complete planning model. In general, simple observer models are easier to maintain aligned with the actual expectations of the user, while those that are more structured allow to simulate with greater detail the inferences of the observer. Also, structured models can be selectively changed through a reconciliation process \cite{chakraborti2017plan} thus ultimately allowing the agent to autonomously re-align its model with the observer's whenever it detects the need.

To the best of our knowledge very little work exists in RL relating to interpretable behavior as we just described. Both \cite{bied2020integrating,zhao2020actor} propose methods relying on a transposition of the original formulation of legibility. The methods result applicable only for goal-driven policies, thus excluding all other types of policies available in various RL frameworks. In addition, they require to specify distance measures between states that, while easy for manipulators working in the cartesian space, can be a difficult task for arbitrary state-spaces. 

Rather than relying of goal locations, we define a legibility criteria that is directly applicable on policies. A regularization method similar to ours is proposed in works on offline policy learning \cite{kostrikov2021offline,wu2019behavior,mysore2021regularizing} where during training the agent's on-policy behavior is regularized towards another behavior. We can see our method as a specific application inside this class methods, where the policy is regularized towards the legible policy.

\section{Method}

The main goal of interpretable behavior is to bring the intention predicted by the observer's model close to the intention of the agent, and to maintain such closeness in time. Consistently with the definition of a legible intention we define a legible policy as: \textit{An agent's policy is legible if it is discernible from a set of other policies}. It is useful to work with this definition because it reflects the general case where an observer is attempting to understand which policy the agent is currently enacting among a set of candidates. Furthermore, the definition doesn't pose constraint on the type of policy but can be applied to arbitrary policies. The goal of legibility is therefore to help the observer to infer the correct policy from the set of those being considered. For this case we hypothesize an observer watching the agent and inferring the policy it is currently pursuing.

\begin{figure}[h]
    \centering

    \begin{tikzpicture}[align=center] 
    \usetikzlibrary{shapes,backgrounds,shapes.misc, positioning,shapes.geometric,arrows,matrix,fit,calc}
    \tikzstyle{main}=[draw, ellipse, minimum height=1cm, minimum width=1cm, align=center]

    \node[main]                         at (0, 0)          (1) {$S$};
    \node[main]                         at (-1.5,0)        (2) {$\Pi$};
    \node[main]                         at (-0.75, -1.5)   (3) {$A$};
    \node[draw=white,text width=2cm]    at (-0.75,1.2) {$P_R(\Pi,S,A)$};
    
    \node[main]                         at (4, 0)         (4) {$\Pi$};
    \node[main]                         at (5.5, 0)       (5) {$S$};
    \node[main]                         at (4.75, -1.5)   (6) {$A$};
    \node[draw=white,text width=2cm]    at (4.75, 1.2) {$P_R^H(\Pi,S,A)$};
    
    \begin{scope}[on background layer]
        \node[draw, thick, rounded corners = 2.5ex, fit=(1) (2) (3),inner sep=3mm, opacity=0.6](Fit1) {};
        \node[draw, thick, rounded corners = 2.5ex, fit=(4) (5) (6),inner sep=3mm, opacity=0.6](Fit2) {};
    \end{scope}

    \draw[->] (1) -- (3);
    \draw[->] (2) -- (3);
    \draw[->] (4) -- (6);
    \draw[->] (5) -- (6);

    \draw[<->] (Fit1) -- node [text width=2cm,midway,above ] {$H(P_R, P_R^H)$} (Fit2);

    \end{tikzpicture}
    \caption{Agent model and second-order theory of mind as equivalent Bayesian Networks. The networks model how agent and observer respectively select and infer actions using the current state and a set of predefined policies, while the function $H$ measures the distance between these two processes.}
    \label{fig:model}
\end{figure}
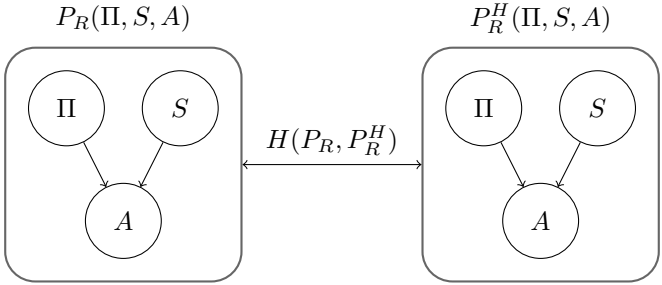

The agent can simulate the presence of an observer by implementing a second-order theory of mind modeling the expectations that it is using to infer intentions. To implement the second-order theory of mind we utilize a middle way between the expressiveness of a complete agent model, and the simplicity of using a hand-crafted solution. This model for theory of mind reasoning, that we refer to as the \textit{Mirror Agent Model}, describes agent and observer models as two equivalent Bayesian networks denoted $P_R$ and $P_R^H$ (Figure~\ref{fig:model}). $P_R$ determines how the agent acts, while $P_R^H$ is the observer's model of how the agent acts. Since the real observer model is part of the observer it is not directly accessible by the agent. The agent must therefore for all computations rely on the estimated model $P_R^H$, the second-order theory of mind. To simplify notations, we make in the following no distinction between these two entities, and we use observer model and second-order theory of mind as interchangeable.

The Bayesian networks are structurally the same and describe the agent as a Markov Decision Process (MDP) with multiple possible policies, however, the random variables ($\Pi, S$ and $A$) can be differently distributed in $P_R$ compared to $P_R^H$, depending on the agent's reasoning and prior information about the observer. A simplifying assumption this model makes is that the user internalizes an agent model with the same structure as the true agent model.   While this assumption may not hold in the general case, it can, for example, be achieved by communicating the agent model, or by performing model alignment dialogues with the goal of communicating the latent variables that the agent uses to act.

We assume that the agent has a fixed set of pre-trained policies identified by the random variable  $\Pi = \{\pi_0,...,\pi_n\}$. Notably, among these there is the currently pursued policy $\pi_R$ with $P_R(\Pi = \pi_R)=1$. Initially, the observer is modelled as ignorant of which policy the agent is pursuing, leading to a uniform prior of the policies: $\forall i\: P_R^H(\pi_i) = k,\: k = \frac{1}{\mid \Pi\mid}$. When using Q-learning, two corresponding Q-value tables $Q_R(a,\pi,s)$ and $Q_R^H(a,\pi,s)$ respectively determine the probability distribution for the agent selecting actions, with $P_R(a\vert \pi, s) = f(Q_R(\pi, s, a))$, and for the observer inferring the agent's actions, with $P_R^H(a\vert \pi, s) = g(Q_R^H(\pi, s, a))$. The Q-value tables can be obtained using any of the available RL methods, while $f$ and $g$ are arbitrary functions that transform Q-values into probability distributions of actions, for example the Boltzmann or the $\epsilon$-greedy distributions \cite{szepesvari2010algorithms}.

To be legible, the agent should select actions that communicate the observer its policy $\pi_R$, while  avoiding communicating the others. This is obtained by selecting actions based on how they reduce the distance between the probability distribution over the agent policies, $P_R(\Pi)$, and the corresponding distribution $P_R^H(\Pi\vert s, a)$ that the observer infers, given an observation of state-action pair. As distance measure we use cross-entropy, obtaining the following formulation:

\begin{align}
    &H(P_R(\Pi), P_R^H(\Pi\vert s,a)) &=\nonumber \\
    &-\log P_R^H(\pi_R\vert a, s) &=\nonumber \\
            &-\log P_R^H(a\vert \pi_R, s) + \log \mathbb{E}[ P_R^H(a\vert \pi, s)] - \log P_R^H(\pi_R). \label{eq:legibility}
\end{align}

Since the action probabilities in Q-learning depend on the Q-values, we can use Eq.~\ref{eq:legibility} to define regularized versions of the Q-values as:
\begin{align}
    &Q_{\text{leg}}(\pi_R, s, a) &=  \nonumber\\
    &Q_R(\pi_R, s, a) - \alpha H(P_R(\Pi), P_R^H(\Pi\vert s,a)) &= \nonumber\\
    &Q_R(\pi_R, s, a) + \alpha \log P_R^H(\pi_R\vert a,s)).&~\label{eq:policy}
\end{align}
\noindent
with $\alpha > 0$ determining the magnitude of regularization. In this way, the right part of Eq.~\ref{eq:policy} regularizes the resulting policy such that the selected actions aim at a small distance between the agent policy and the policy inferred by the observer. Equation~\ref{eq:legibility} expresses that the resulting decision boundary introduced by legibility impacts the states in which $\pi_R(s)$ returns an action that has high probability also in other policies. In these cases, a trade-off between such actions, and a sub-obtimal/legible actions is made.

\section{Experiments and evaluation}

We tested and evaluated the proposed model with two experiments. The first is an illustrative example in a gridworld setting and is intended to provide insight into how the legible policy modifies the original policy. The second experiment  is more extensive and is  performed with a Deep Q-Network.

\subsection{Grid-world experiment}

In this experiment we tested the proposed method on a gridworld scenario. The grid is 7x7 and without obstacles. There are 3 possible goals at the corners, for which we trained three corresponding policies with Q-learning. For simplicity we set $Q_R = Q_R^H$ and $f=g$, meaning that the agent assumes the observer to use the same Q-values and derived action probabilities as its own, i.e., $\forall i\: P_R(A|\pi_i,S) = P_R^H(A|\pi_i, S)$. $\alpha$ was set to 1,  which has the advantage of not require modeling how the observer models the task, which is a costly procedure. However, nothing prohibits usage of different Q-values for the observer. In such cases, the agent would be evaluated by a different set of policies than those it possesses. 

\begin{figure}
\centering
\begin{tabular}{m{5pt}m{.22\textwidth}m{.22\textwidth}}
  & Original & Legible\\
  $g_0$ & \includegraphics[width=.20\textwidth]{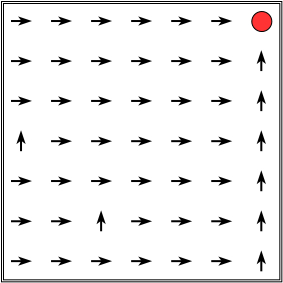} &   \includegraphics[width=.20\textwidth]{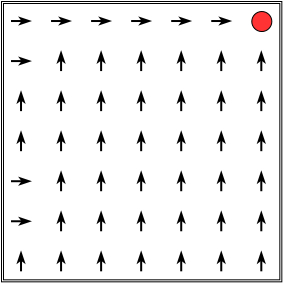} \\
  $g_1$ & \includegraphics[width=.20\textwidth]{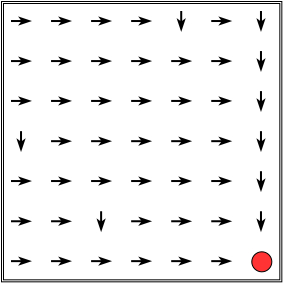} &   \includegraphics[width=.20\textwidth]{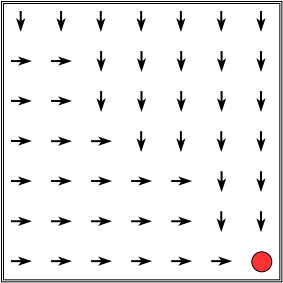} \\
  $g_2$ & \includegraphics[width=.20\textwidth]{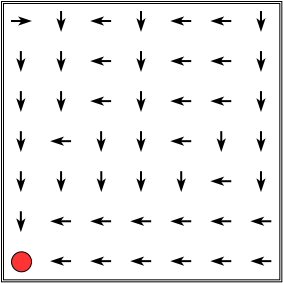} &   \includegraphics[width=.20\textwidth]{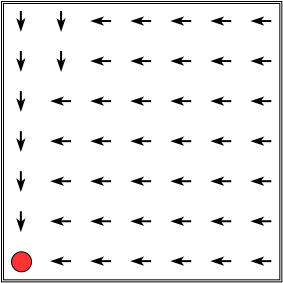} \\

\end{tabular}
\caption{Left: policies for the three goals (red dots) learned with Q-learning. Right: legible policies. The legible policies avoid ambiguity of goal location.}
\label{fig:results}
\end{figure}

Figure~\ref{fig:results} shows in the left column the optimal policies learned by the agent. In the right column the correspondingly legible policies obtained using $\alpha=1$. The learned policies move towards a wall adjacent the goal, and then approach the goal by walking along the wall. However, to be legible, it is important to approach the right wall that disambiguates the goal location. The legible policies systematically approach an unambiguous wall. Notice also how for $g_1$, the legible policy makes the agent walk in the middle to avoid approaching the other goals.

\subsection{Deep Q-Network experiment}

\begin{figure}[h]
\centering
    \resizebox{.8\linewidth}{!}{
    \includegraphics[]{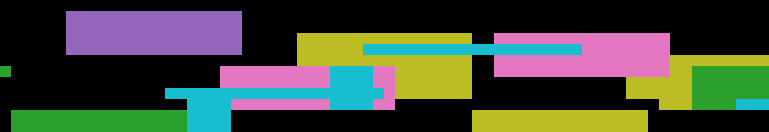}
    }
    \caption{Sampled tunnel environment. While traversing a tunnel the agent is rewarded to walk on cells of its same color (green). Hitting an obstacle (teal) instead punishes the agent and resets the episode.}
    \label{fig:tunnel}
\end{figure}

In the second experiment we used \emph{OpenAI Gym} \cite{brockman2016openai}. We designed a  simulated environment in which the agent had to pass through tunnels of length $L$ and width $W$, composed of $C+2$ types of cells: empty cells, obstacle cells, and $C$ types of cells of different colors (see Figure~\ref{fig:tunnel}). The agent was defined to see a maximum of $S$ cells in front of it and had 3 possible actions: move one cell up, move one cell down, or stay at the same position. If the agent moves to a colored cell it receives a reward of $+1$ while if it moves to an obstacle it gets a punishment of $-10$ and a new episode restarts. Moving to an empty cell or to a cell of a color different from its own does not result in any reward or punishment. The environment is not goal-oriented but rather defines regions of reward and of punishment for the agent. These regions can be of arbitrary shape and we used rectangles for colored regions and squares or lines for obstacles.

Since the agent is unaffected by cells of a color different from its rewarding color, to simplify the learning process it was trained on tunnels containing only one color and obstacles. Later, tunnels containing $C$ colors are obtained by using $C$ tunnels sharing obstacles and agent position. Inside a single-color tunnel, at every timestep the observation corresponds to a set of three matrices $M_0, M_1, M_2$ of size $W\cdot S$, each representing a slice of the tunnel up to the agent's sight distance. The first matrix contains only colored cells, the second only obstacles and the third the agent's position. Inside every matrix, each cell is characterized by the summation of three embedding vectors:
\begin{align}
    c_{ij} = w_i + s_j + t_{ij}.
\end{align}

\noindent where $w_i$ and $s_j$ are position embeddings identifying the cell inside the matrix. For example, $\langle w_0, s_5 \rangle$ indicates cell $0-5$. While $t_{ij}$ identifies whether that cell is occupied: in $M_0$ a cell is occupied if it is colored, in $M_1$ if it is an obstacle, in $M_2$ if it contains the agent's position.
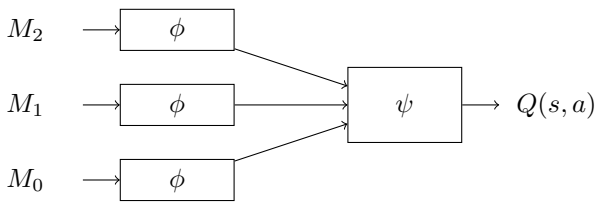
\begin{figure}[h]
    \centering

    \begin{tikzpicture}[align=center] 
    \tikzstyle{input}=[minimum height=1cm, minimum width=1.5cm, align=center]
    \tikzstyle{main}=[draw, minimum width=1.5cm, align=center]
    
    \node[input]                        at (0, 0)   (1) {$M_0$};
    \node[input]                        at (0, 1)   (2) {$M_1$};
    \node[input]                        at (0, 2)   (3) {$M_2$};
    
    \node[main]                        at (2, 0)   (4) {$\phi$};
    \node[main]                        at (2, 1)   (5) {$\phi$};
    \node[main]                        at (2, 2)   (6) {$\phi$};
    
    \node[main, minimum height=1cm]    at (5, 1)   (7) {$\psi$};
    \node[input]                        at (7, 1)   (8) {$Q(s, a)$};
    
    \draw[->] (1) -- (4);
    \draw[->] (2) -- (5);
    \draw[->] (3) -- (6);
    
    \draw[->] (4) -- (7);
    \draw[->] (5) -- (7);
    \draw[->] (6) -- (7);
    
    \draw[->] (7) -- (8);
    
    \end{tikzpicture}
    
    \caption{Q-network for the tunnel enviroment. $\phi$: convolution network shared by the three inputs. $\psi$: fully-connected network}
    
    \label{fig:tunnel_network}
    
\end{figure}

Figure~\ref{fig:tunnel_network} shows the employed Q-network. In the network, $\phi$ is a convolution network which convolves on the matrices of embeddings, and is shared by all the inputs $M_0, M_1$ and $M_2$. $\psi$ is a fully connected network that takes as input the vector $\langle \phi (M_0), \phi (M_1), \phi (M_2) \rangle$ and outputs a vector of size 3 for the Q-values.

We trained the agent on 30000 random, single-color tunnels of length 200 and width 12 cells, while the agent's observation windows was set to 20 cells. For every tunnel 5 colored rectangles and 10 obstacles of shape square or line were randomly placed. Table~\ref{tab:tunnel_hyperparams} shows the network's hyperparameters used for training the Q-network.

\renewcommand{\arraystretch}{1.5}
\begin{table}[h]
\centering
    \resizebox{.5\linewidth}{!}{
    \begin{tabular}{lll}
        \hline
        \textbf{Parameter} & \textbf{Amount} & \textbf{Type} \\
        \hline
        $\text{Layers}_{\phi}$ & (100,100,100,100) & $2D$-convolution\\
        \hline
        $\text{Layers}_{\psi}$ & (200,200,50) & Fully-connected\\
        \hline
         Embedding size& 100 & \\
        \hline
         Learning rate& $1e-3$ & \\
        \hline
        Learning rate decay & $1e-4$ & \\
        \hline
        Buffer size & 150000 & \\
        \hline
        Policy $\epsilon$ & 0.1 & \\
        \hline
        Discount factor & 0.98 & \\
        \hline
    \end{tabular}
    }
    \caption{Hyperparameters of the Q-network for training the agent.}
    \label{tab:tunnel_hyperparams}
\end{table}

As previously mentioned, after training to obtain a tunnel with $C$ colors we merged $C$ tunnels at once, with each tunnel containing only cells of the respective color, while all sharing the same obstacles and agent position. In this way, at each step the agent has $C$ different policies to follow, each one seeking a particular color. This is equal to the result of training $C$ different policies simultaneously.

\subsubsection{Quantitative Evaluation}

We tested the proposed method for legible policy in a setting where both agent and observer  use the same Q-function (the trained Q-network) and the greedy policy to always select the action with highest Q-value. Since the introduced regularization penalizes actions with high probability in other policies, we expected the agent to avoid cells of colors that are not its own. In other words, since the observer model judges the agent's behavior by confronting it with policies that seek cells of given colors, by avoiding cells of other colors the agent decreases the probability of those policies in the observer's inferences.

We tested this hypothesis first quantitatively by measuring the average gathered reward  over 200 episodes, while using increasing values for the regularization factor $\alpha$. Every random tunnel had $C=4$ colors, 5 rectangular colored patches for each color, and 10 square obstacles. In this setting we measured the reward gathered by the agent when pursuing the color $C_0$, and the average reward for the other colors $C_{1..3}$ accumulated while pursuing $C_0$. We then divided these scores by the maximum rewards that the policy could have gathered for the corresponding colors, thereby obtaining a \emph{reward ratio} with values between 0 and 1. For example, a reward ratio of 0.5 means  that the agent accumulated half of the possible maximal reward. As a complement to the reward ratio,  \emph{success rate} was computed as the probability of succeeding, i.e.  reaching the end of the tunnel without hitting any obstacles during an episode. Table~\ref{tab:tunnel_res} summarizes this experiment.

\renewcommand{\arraystretch}{1.5}
\begin{table}[h]
    \centering
    \resizebox{.5\linewidth}{!}{
    \begin{tabular}{|c|c|c|c|c|c|c|c|}
    \hline
         & $\alpha=0$ & $\alpha=0.1$ & $\alpha=0.5$ & $\alpha=1$ & $\alpha=2$ & $\alpha=5$  \\ \hline
        $C_0$  & 0.8 & 0.8 & 0.76 & 0.76 & 0.77 & 0.75 \\ \hline
        $C_{1..3}$ & 0.29 & 0.21& 0.15 & 0.14 & 0.13 & 0.11 \\ \hline
        Success & 0.99 & 0.95 & 0.96 & 0.95 & 0.92 & 0.87\\ \hline
    \end{tabular}
    }
    \caption{Average accumulated reward ration by the policies for color $C_0$ and colors $C_{1..3}$ for increasing values of $\alpha$. The row \textit{Success} indicates the probability of completing a tunnel without hitting obstacles.}
    \label{tab:tunnel_res}
\end{table}

Table~\ref{tab:legibility} instead summarizes the degree of legibility of the agent's policy measured as the expected probability that the observer model gives to the agent's policy through the episodes:
\begin{align}
    \mathcal{L} = \underset{\langle a,s \rangle \sim \mathcal{E}}{\mathrm{E}} [P_R^H(\pi_R|s,a)].
\end{align}
\noindent
where every state transition is given equal weight $p(s,a) = \frac{1}{|\mathcal{E}|}$. The second row of the table shows the gain of legibility obtained by using the legible policy rather that the original:
\begin{align}
    \mathcal{L}_{\text{gain}}(x) = \frac{\mathcal{L}_{\alpha=x}}{\mathcal{L}_{\alpha=0}}.
\end{align}

We tested and evaluated the proposed model with two experiments. The first is an illustrative example in a gridworld setting and is intended to provide insight into how the legible policy modifies the original policy. The second experiment  is more extensive and is  performed with a Deep Q-Network.

\renewcommand{\arraystretch}{1.5}
\begin{table}[h]
\centering
    \resizebox{.5\linewidth}{!}{
    \begin{tabular}{|c|c|c|c|c|c|c|c|}
    \hline
          & $\alpha=0$ & $\alpha=0.1$ & $\alpha=0.5$ & $\alpha=1$ & $\alpha=2$ & $\alpha=5$ \\ \hline
         $\mathcal{L}$ & 0.30 & 0.36 & 0.44 & 0.48 & 0.48 & 0.51  \\ \hline
         $\mathcal{L}_{\text{gain}}$ & 1 & 1.2 & 1.46 & 1.6 & 1.6 & 1.7 \\ \hline
    \end{tabular}
    }
    \caption{Policy legibity for increasing values of $\alpha$.}
    \label{tab:legibility}
\end{table}

\subsubsection{Qualitative Evaluation}

\begin{figure}
   \centering
    \begin{tabular}{m{0.2cm}cc}
    &$\alpha=0$ & $\alpha=1$\\
    1) & \includegraphics[width=5.2cm]{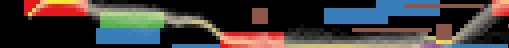}&
    \includegraphics[width=5.2cm]{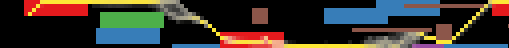}\\
    2) & \includegraphics[width=5.2cm]{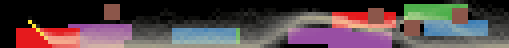}&
    \vspace{1mm}\includegraphics[width=5.2cm]{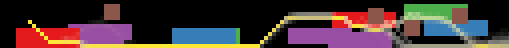}\\
    \end{tabular}
    \caption{Qualitative results for two sampled tunnels. The left column shows the agent's optimal learned behavior while the left side its regularized version. }
    \label{fig:qualitative_results_tunnels}
\end{figure}

\begin{figure}
   \centering
    \resizebox{\linewidth}{!}{
    \begin{tabular}{m{0.2cm}cccccc}
    & $\alpha=0$ & $\alpha=0.1$ & $\alpha=0.5$ & $\alpha=1$ & $\alpha=2$ & $\alpha=5$\\
    
    1) & \includegraphics[width=2.5cm,trim=50 100 50 100,clip]{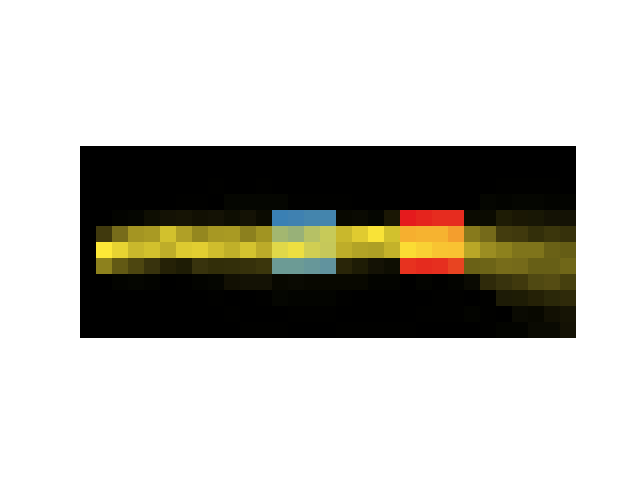}&
    \includegraphics[width=2.5cm,trim=50 100 50 100,clip]{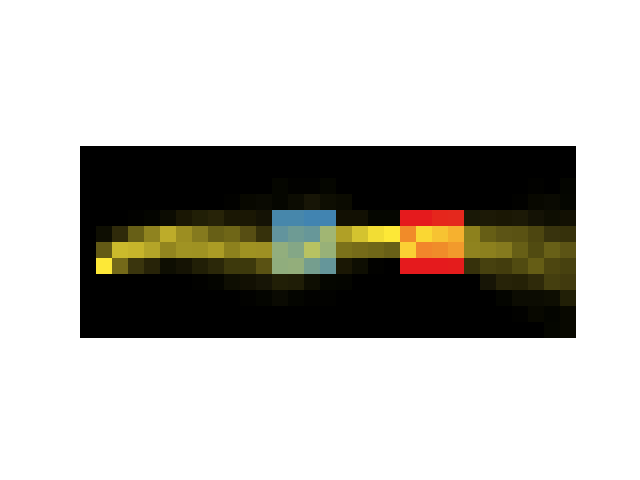}&
    \includegraphics[width=2.5cm,trim=50 100 50 100,clip]{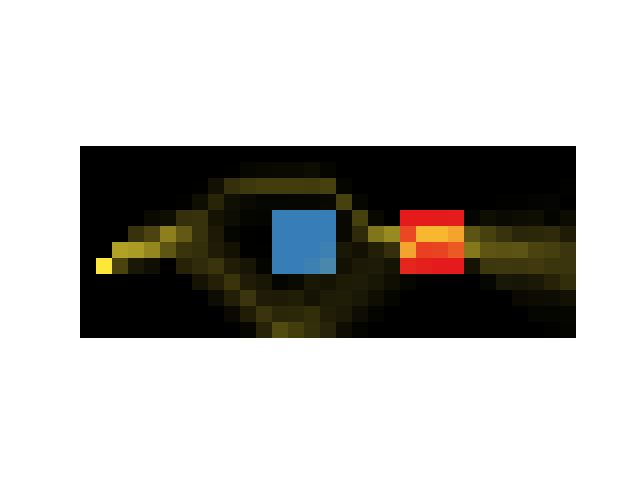}&
    \includegraphics[width=2.5cm,trim=50 100 50 100,clip]{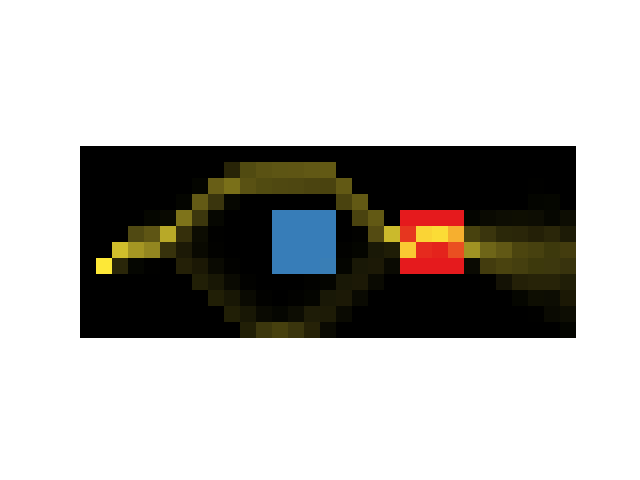}&
    \includegraphics[width=2.5cm,trim=50 100 50 100,clip]{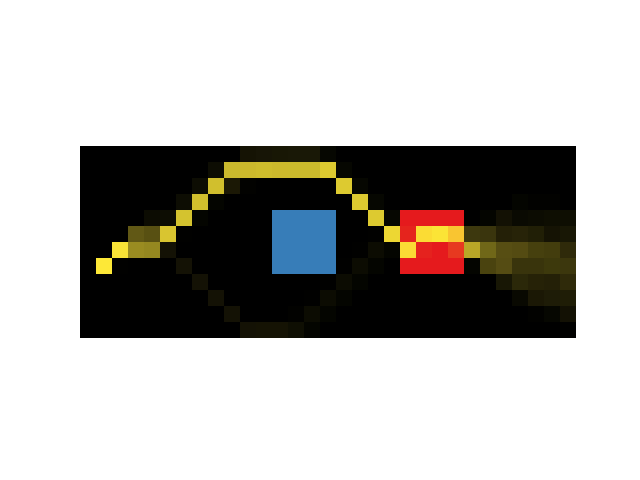}&
    \includegraphics[width=2.5cm,trim=50 100 50 100,clip]{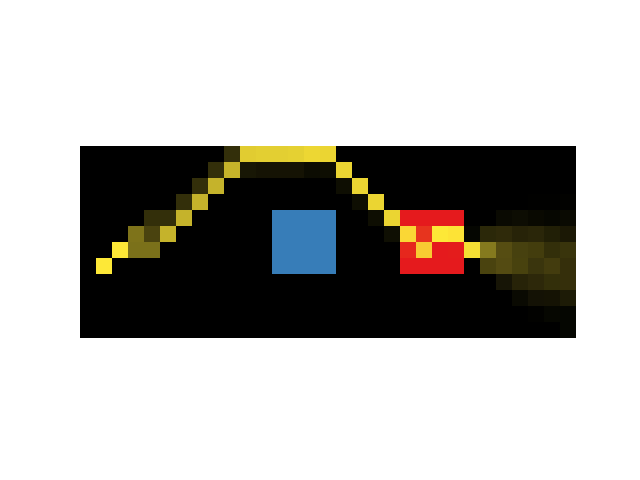}\\
    2) & \includegraphics[width=2.5cm,trim=50 100 50 100,clip]{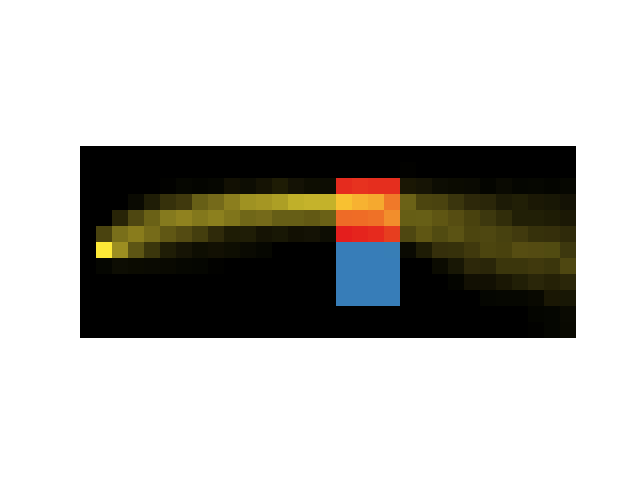}&
    \includegraphics[width=2.5cm,trim=50 100 50 100,clip]{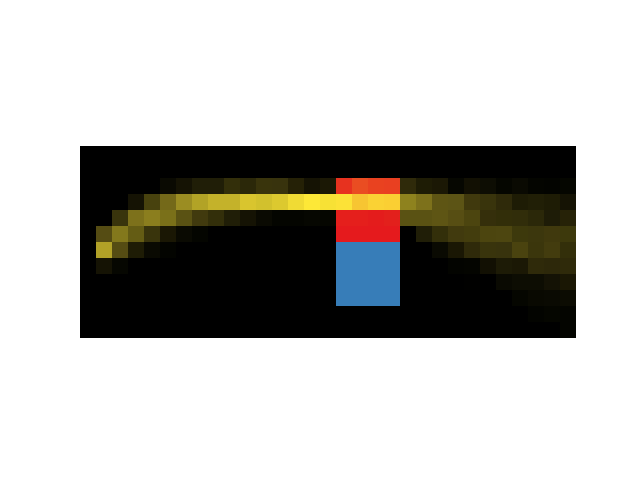}&
    \includegraphics[width=2.5cm,trim=50 100 50 100,clip]{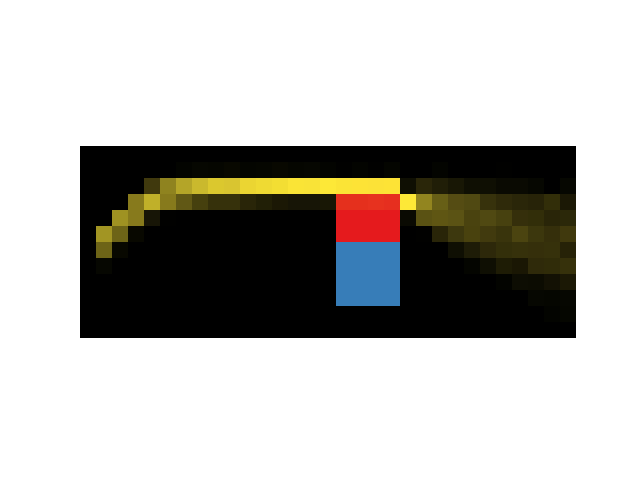}&
    \includegraphics[width=2.5cm,trim=50 100 50 100,clip]{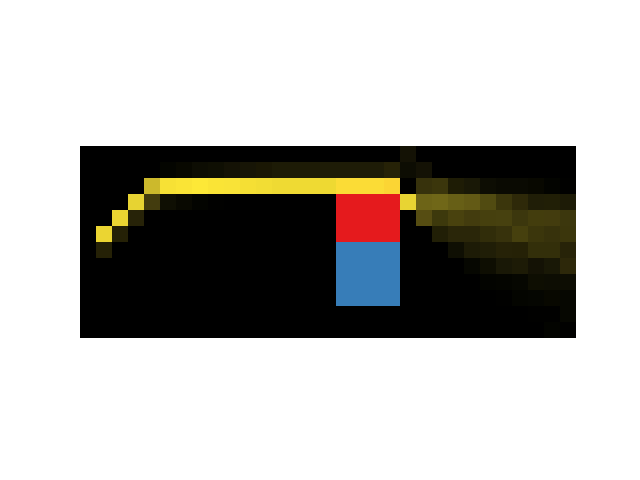}&
    \includegraphics[width=2.5cm,trim=50 100 50 100,clip]{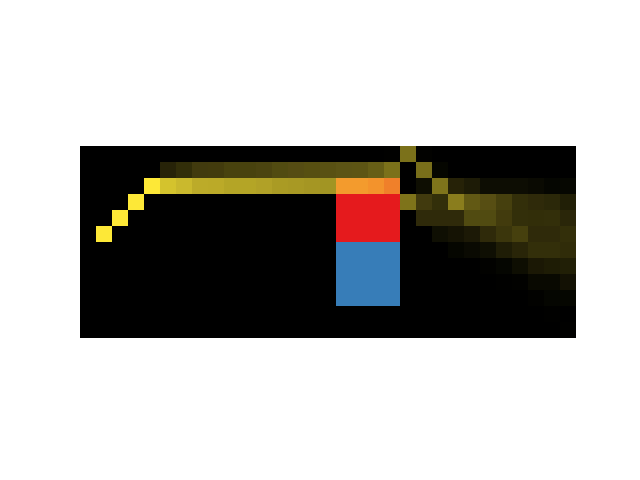}&
    \includegraphics[width=2.5cm,trim=50 100 50 100,clip]{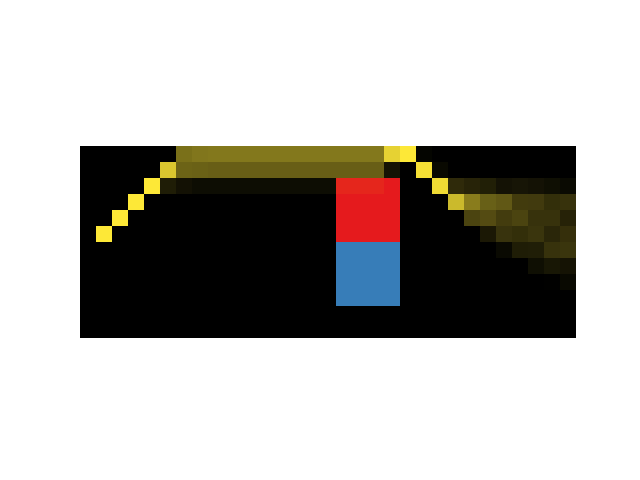}\\
    3) & \includegraphics[width=2.5cm,trim=50 100 50 100,clip]{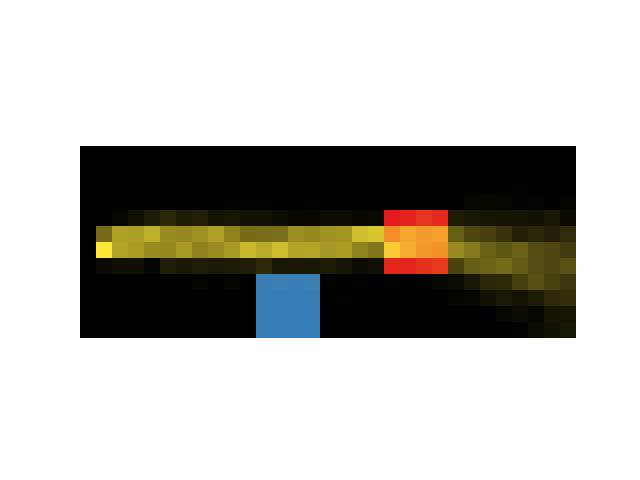}&
    \includegraphics[width=2.5cm,trim=50 100 50 100,clip]{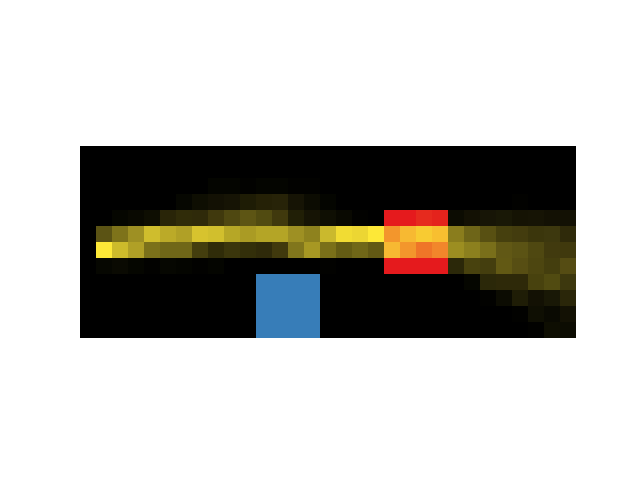}&
    \includegraphics[width=2.5cm,trim=50 100 50 100,clip]{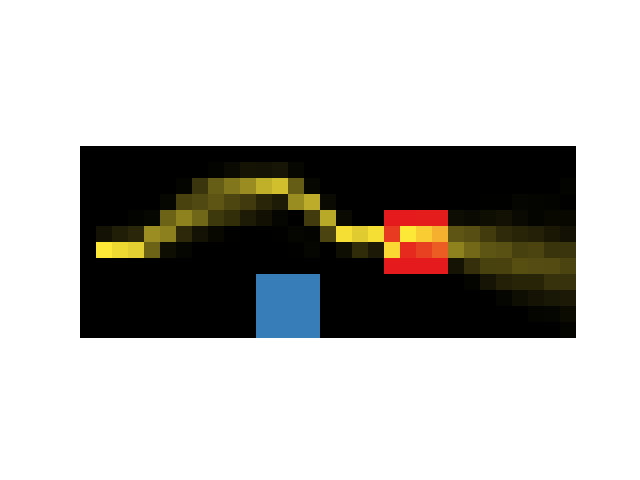}&
    \includegraphics[width=2.5cm,trim=50 100 50 100,clip]{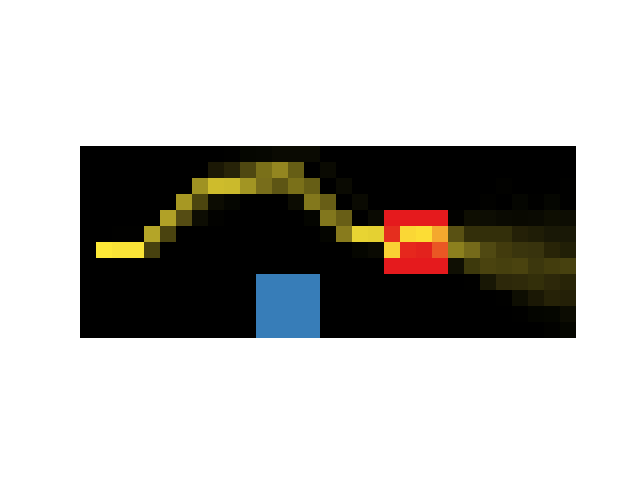}&
    \includegraphics[width=2.5cm,trim=50 100 50 100,clip]{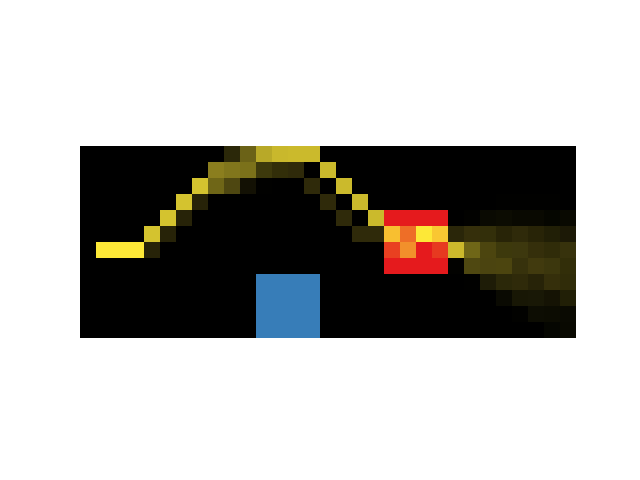}&
    \includegraphics[width=2.5cm,trim=50 100 50 100,clip]{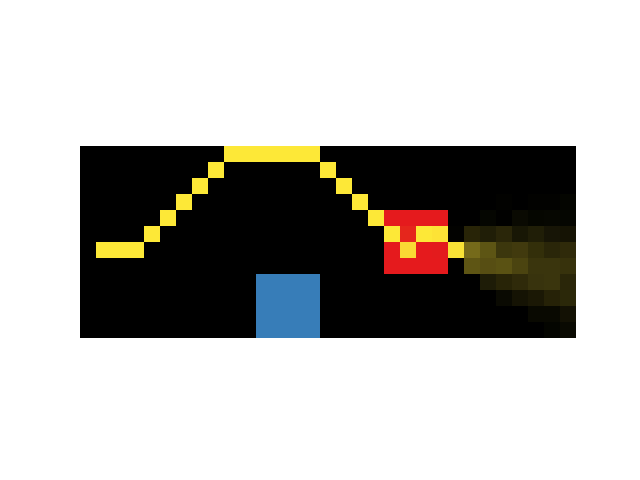}\\
    \end{tabular}
    }
    \caption{Qualitative results for three types of positions of reward regions and increasing levels of $\alpha$.}
    \label{fig:qualitative_results}
\end{figure}

Figure~\ref{fig:qualitative_results_tunnels} shows the effect of regularization on two sampled tunnels. In the plots red is the rewarding color of the agent and obstacles are in brown. The trajectories in yellow are obtained by simulating and averaging 200 trials. In addition, to better understand the effect of regularization, the agent's behaviors for three different configurations of colored regions and increasing factor $\alpha$ are plotted in  Figure~\ref{fig:qualitative_results}.  The plots  have two colors: red as reward color for the agent's policy, and blue as reward color for a different policy.

Legibility clearly skews the trajectories such that they pass farther away from non-red cells in a way that is proportional to $\alpha$. Notice also how in Figure~\ref{fig:qualitative_results} regularization becomes detrimental for values of $\alpha$ that are too high. In such cases, the agent's original policy of walking over red cell is dominated by the regularization to avoid blue cells, and in some cases the agent is not able to pass over any red cells even if there aren't any obstacles.

\section{Discussion}

Our quantitative results indicate that the proposed regularization for legibility is effective in making the observer model discriminate the true agent's policy. This is highlighted in Table~\ref{tab:tunnel_res} where we can see that the reward ratio for colors different from $C_0$ decrease as $\alpha$ increases, signifying that the agent avoids regions with colors different from its own. The qualitative results also confirm this observation, by showing that as $\alpha$ increases so does the effort of the regularized policy to avoid other colors. 

We calculated how high values of $\alpha$ are detrimental both in terms of accumulated reward and success rate of the episodes. The reason is that the agent is regularized so much to avoid other  policies that its original policy is overridden rather than regularized. However, the agent incurs a noticeable loss in terms of accumulated reward or success rate only for high regularization factors.

The general results confirm our originally formulated hypothesis that legibility increases the probability of the agent policy appearing in the observer's model by making the agent avoid rewarding regions of other policies. This is an emergent behavior that was not coded in the equations and in our experiments represented a generalization over goal-driven solutions for legibility, because the reward regions of goal-driven policies are a special case of those of arbitrary policies that put reward at goal locations.

Furthermore, we noticed that our results are obtained with a simple observer model having uniformly distributed probabilities, and that considers the same policies that are available to the agent. Because our results are qualitative similar to those reported in earlier works on legibility \cite{dragan2013legibility,bied2020integrating} ie. legible trajectories are skewed to avoid other goal locations, it is suggested that a similarly uniformly initialized observer model was implicitly utilized in those papers as well. However, in contrast with previous solutions the proposed method allows to easily generalize over the difference between agent and observer models, that is by using different corresponding networks. This was not possible in previous methods because the observer was fixed.

\section{Conclusion}

In this paper we introduced a model that allows to regularize a reinforcement learning agent for legibility.  In our formulation we propose a legibility criteria that induces an observer model to disambiguate the agent's intention from a set of others, with intentions being implemented as policies. We suggest that rather than modifying the learning procedure of the agent we can wrap a priorly learned set of policies by a pair of Bayesian Networks that model agent and observer respectively. The coupled networks describes a setting of second-order theory of mind that, by reasoning on how the observer infers policies, increases the discrimination between the agent's true policy and other candidate policies. 

We evaluated the method on an illustrative example showing how legibility impacts the decision boundary of the agent, and on a Deep-RL example. In general, our model is successful at increasing the legibility of trajectories without incurring in losses for the agent when the regularization factor is kept at a reasonable level. Furthermore, our qualitative results show that the obtained trajectories are similarly arced as those obtained in earlier work on Explainable Planning, but with the main difference of computing legibility on reward regions rather goal states.

The proposed methods introduces two relevant degrees of freedom in legibility. The first is that legibility is computed with respect to reward regions rather than goal locations. This allows to regularize arbitrary policies and especially those that can run indefinitely. Policies of this type can't be regularized by methods relying on the original formulation of legibility because of the need of a goal state. The second degree of freedom is on the possibility of decoupling agent and observer models. This allows to specify that the the observer uses a different reward distribution, and legibility is to be computed against that distribution rather than the agent's. This decoupling is not easy to implement using previous methods relying on distance measures, because would require to specify how the observer measures distances on the state-space.

Since the agent's learning algorithm is unmodified, it is straightforward to apply our method to arbitrary problems and types of agents. Even though we couldn't test it on extensive test beds of agents and problems it is reasonable to think that problems effectively captured as MDPs can be regularized without major additional implementations.

\section{Declarations}

\begin{itemize}
    \item The author declares that there are no conflicts of interest associated with this study.
    \item  This study did not involve human participants or animals.
    \item This study was funded by the Department of Computing Science of Ume{\aa} University, Universitetstorget 4, Ume{\aa}, 901 87, Sweden.
\end{itemize}

\bibliography{bibliography}


\begin{thebibliography}{32}
\ifx \bisbn   \undefined \def \bisbn  #1{ISBN #1}\fi
\ifx \binits  \undefined \def \binits#1{#1}\fi
\ifx \bauthor  \undefined \def \bauthor#1{#1}\fi
\ifx \batitle  \undefined \def \batitle#1{#1}\fi
\ifx \bjtitle  \undefined \def \bjtitle#1{#1}\fi
\ifx \bvolume  \undefined \def \bvolume#1{\textbf{#1}}\fi
\ifx \byear  \undefined \def \byear#1{#1}\fi
\ifx \bissue  \undefined \def \bissue#1{#1}\fi
\ifx \bfpage  \undefined \def \bfpage#1{#1}\fi
\ifx \blpage  \undefined \def \blpage #1{#1}\fi
\ifx \burl  \undefined \def \burl#1{\textsf{#1}}\fi
\ifx \doiurl  \undefined \def \doiurl#1{\url{https://doi.org/#1}}\fi
\ifx \betal  \undefined \def \betal{\textit{et al.}}\fi
\ifx \binstitute  \undefined \def \binstitute#1{#1}\fi
\ifx \binstitutionaled  \undefined \def \binstitutionaled#1{#1}\fi
\ifx \bctitle  \undefined \def \bctitle#1{#1}\fi
\ifx \beditor  \undefined \def \beditor#1{#1}\fi
\ifx \bpublisher  \undefined \def \bpublisher#1{#1}\fi
\ifx \bbtitle  \undefined \def \bbtitle#1{#1}\fi
\ifx \bedition  \undefined \def \bedition#1{#1}\fi
\ifx \bseriesno  \undefined \def \bseriesno#1{#1}\fi
\ifx \blocation  \undefined \def \blocation#1{#1}\fi
\ifx \bsertitle  \undefined \def \bsertitle#1{#1}\fi
\ifx \bsnm \undefined \def \bsnm#1{#1}\fi
\ifx \bsuffix \undefined \def \bsuffix#1{#1}\fi
\ifx \bparticle \undefined \def \bparticle#1{#1}\fi
\ifx \barticle \undefined \def \barticle#1{#1}\fi
\bibcommenthead
\ifx \bconfdate \undefined \def \bconfdate #1{#1}\fi
\ifx \botherref \undefined \def \botherref #1{#1}\fi
\ifx \url \undefined \def \url#1{\textsf{#1}}\fi
\ifx \bchapter \undefined \def \bchapter#1{#1}\fi
\ifx \bbook \undefined \def \bbook#1{#1}\fi
\ifx \bcomment \undefined \def \bcomment#1{#1}\fi
\ifx \oauthor \undefined \def \oauthor#1{#1}\fi
\ifx \citeauthoryear \undefined \def \citeauthoryear#1{#1}\fi
\ifx \endbibitem  \undefined \def \endbibitem {}\fi
\ifx \bconflocation  \undefined \def \bconflocation#1{#1}\fi
\ifx \arxivurl  \undefined \def \arxivurl#1{\textsf{#1}}\fi
\csname PreBibitemsHook\endcsname

%%% 1
\bibitem{Sule19}
\begin{bchapter}
\bauthor{\bsnm{Anjomshoae}, \binits{S.}},
\bauthor{\bsnm{Najjar}, \binits{A.}},
\bauthor{\bsnm{Calvaresi}, \binits{D.}},
\bauthor{\bsnm{Fr\"{a}mling}, \binits{K.}}:
\bctitle{Explainable agents and robots: Results from a systematic literature
  review}.
In: \bbtitle{Proceedings of the 18th International Conference on Autonomous
  Agents and MultiAgent Systems}.
\bsertitle{AAMAS '19},
pp. \bfpage{1078}--\blpage{1088}.
\bpublisher{International Foundation for Autonomous Agents and Multiagent
  Systems},
\blocation{Richland, SC}
(\byear{2019})
\end{bchapter}
\endbibitem

%%% 2
\bibitem{hellstrom2018understandable}
\begin{barticle}
\bauthor{\bsnm{Hellstr{\"o}m}, \binits{T.}},
\bauthor{\bsnm{Bensch}, \binits{S.}}:
\batitle{Understandable robots-what, why, and how}.
\bjtitle{Paladyn, Journal of Behavioral Robotics}
\bvolume{9}(\bissue{1}),
\bfpage{110}--\blpage{123}
(\byear{2018})
\end{barticle}
\endbibitem

%%% 3
\bibitem{schaefer2017communicating}
\begin{barticle}
\bauthor{\bsnm{Schaefer}, \binits{K.E.}},
\bauthor{\bsnm{Straub}, \binits{E.R.}},
\bauthor{\bsnm{Chen}, \binits{J.Y.}},
\bauthor{\bsnm{Putney}, \binits{J.}},
\bauthor{\bsnm{Evans~III}, \binits{A.W.}}:
\batitle{Communicating intent to develop shared situation awareness and
  engender trust in human-agent teams}.
\bjtitle{Cognitive Systems Research}
\bvolume{46},
\bfpage{26}--\blpage{39}
(\byear{2017})
\end{barticle}
\endbibitem

%%% 4
\bibitem{chang2018effects}
\begin{bchapter}
\bauthor{\bsnm{Chang}, \binits{M.L.}},
\bauthor{\bsnm{Gutierrez}, \binits{R.A.}},
\bauthor{\bsnm{Khante}, \binits{P.}},
\bauthor{\bsnm{Short}, \binits{E.S.}},
\bauthor{\bsnm{Thomaz}, \binits{A.L.}}:
\bctitle{Effects of integrated intent recognition and communication on
  human-robot collaboration}.
In: \bbtitle{2018 IEEE/RSJ International Conference on Intelligent Robots and
  Systems (IROS)},
pp. \bfpage{3381}--\blpage{3386}
(\byear{2018}).
\bcomment{IEEE}
\end{bchapter}
\endbibitem

%%% 5
\bibitem{sciutti2018humanizing}
\begin{barticle}
\bauthor{\bsnm{Sciutti}, \binits{A.}},
\bauthor{\bsnm{Mara}, \binits{M.}},
\bauthor{\bsnm{Tagliasco}, \binits{V.}},
\bauthor{\bsnm{Sandini}, \binits{G.}}:
\batitle{Humanizing human-robot interaction: On the importance of mutual
  understanding}.
\bjtitle{IEEE Technology and Society Magazine}
\bvolume{37}(\bissue{1}),
\bfpage{22}--\blpage{29}
(\byear{2018})
\end{barticle}
\endbibitem

%%% 6
\bibitem{losey2018review}
\begin{botherref}
\oauthor{\bsnm{Losey}, \binits{D.P.}},
\oauthor{\bsnm{McDonald}, \binits{C.G.}},
\oauthor{\bsnm{Battaglia}, \binits{E.}},
\oauthor{\bsnm{O'Malley}, \binits{M.K.}}:
A review of intent detection, arbitration, and communication aspects of shared
  control for physical human--robot interaction.
Applied Mechanics Reviews
\textbf{70}(1)
(2018)
\end{botherref}
\endbibitem

%%% 7
\bibitem{verberne2012trust}
\begin{barticle}
\bauthor{\bsnm{Verberne}, \binits{F.M.}},
\bauthor{\bsnm{Ham}, \binits{J.}},
\bauthor{\bsnm{Midden}, \binits{C.J.}}:
\batitle{Trust in smart systems: Sharing driving goals and giving information
  to increase trustworthiness and acceptability of smart systems in cars}.
\bjtitle{Human factors}
\bvolume{54}(\bissue{5}),
\bfpage{799}--\blpage{810}
(\byear{2012})
\end{barticle}
\endbibitem

%%% 8
\bibitem{walker2018communicating}
\begin{bchapter}
\bauthor{\bsnm{Walker}, \binits{M.}},
\bauthor{\bsnm{Hedayati}, \binits{H.}},
\bauthor{\bsnm{Lee}, \binits{J.}},
\bauthor{\bsnm{Szafir}, \binits{D.}}:
\bctitle{Communicating robot motion intent with augmented reality}.
In: \bbtitle{Proceedings of the 2018 ACM/IEEE International Conference on
  Human-Robot Interaction},
pp. \bfpage{316}--\blpage{324}
(\byear{2018})
\end{bchapter}
\endbibitem

%%% 9
\bibitem{kulkarni2019explicable}
\begin{bchapter}
\bauthor{\bsnm{Kulkarni}, \binits{A.}},
\bauthor{\bsnm{Zha}, \binits{Y.}},
\bauthor{\bsnm{Chakraborti}, \binits{T.}},
\bauthor{\bsnm{Vadlamudi}, \binits{S.G.}},
\bauthor{\bsnm{Zhang}, \binits{Y.}},
\bauthor{\bsnm{Kambhampati}, \binits{S.}}:
\bctitle{Explicable planning as minimizing distance from expected behavior}.
In: \bbtitle{AAMAS},
pp. \bfpage{2075}--\blpage{2077}
(\byear{2019})
\end{bchapter}
\endbibitem

%%% 10
\bibitem{zhang2017plan}
\begin{bchapter}
\bauthor{\bsnm{Zhang}, \binits{Y.}},
\bauthor{\bsnm{Sreedharan}, \binits{S.}},
\bauthor{\bsnm{Kulkarni}, \binits{A.}},
\bauthor{\bsnm{Chakraborti}, \binits{T.}},
\bauthor{\bsnm{Zhuo}, \binits{H.H.}},
\bauthor{\bsnm{Kambhampati}, \binits{S.}}:
\bctitle{Plan explicability and predictability for robot task planning}.
In: \bbtitle{2017 IEEE International Conference on Robotics and Automation
  (ICRA)},
pp. \bfpage{1313}--\blpage{1320}
(\byear{2017}).
\bcomment{IEEE}
\end{bchapter}
\endbibitem

%%% 11
\bibitem{dragan2013legibility}
\begin{bchapter}
\bauthor{\bsnm{Dragan}, \binits{A.D.}},
\bauthor{\bsnm{Lee}, \binits{K.C.}},
\bauthor{\bsnm{Srinivasa}, \binits{S.S.}}:
\bctitle{Legibility and predictability of robot motion}.
In: \bbtitle{2013 8th ACM/IEEE International Conference on Human-Robot
  Interaction (HRI)},
pp. \bfpage{301}--\blpage{308}
(\byear{2013}).
\bcomment{IEEE}
\end{bchapter}
\endbibitem

%%% 12
\bibitem{persiani2021probabilistic}
\begin{bchapter}
\bauthor{\bsnm{Persiani}, \binits{M.}},
\bauthor{\bsnm{Hellstr{\"o}m}, \binits{T.}}:
\bctitle{Probabilistic plan legibility with off-the-shelf planners}.
In: \bbtitle{9th ICAPS Workshop on Planning and Robotics. ICAPS 2021.}
(\byear{2021})
\end{bchapter}
\endbibitem

%%% 13
\bibitem{puiutta2020explainable}
\begin{bchapter}
\bauthor{\bsnm{Puiutta}, \binits{E.}},
\bauthor{\bsnm{Veith}, \binits{E.M.}}:
\bctitle{Explainable reinforcement learning: A survey}.
In: \bbtitle{International Cross-Domain Conference for Machine Learning and
  Knowledge Extraction},
pp. \bfpage{77}--\blpage{95}
(\byear{2020}).
\bcomment{Springer}
\end{bchapter}
\endbibitem

%%% 14
\bibitem{alharin2020reinforcement}
\begin{barticle}
\bauthor{\bsnm{Alharin}, \binits{A.}},
\bauthor{\bsnm{Doan}, \binits{T.-N.}},
\bauthor{\bsnm{Sartipi}, \binits{M.}}:
\batitle{Reinforcement learning interpretation methods: A survey}.
\bjtitle{IEEE Access}
\bvolume{8},
\bfpage{171058}--\blpage{171077}
(\byear{2020})
\end{barticle}
\endbibitem

%%% 15
\bibitem{du2019techniques}
\begin{barticle}
\bauthor{\bsnm{Du}, \binits{M.}},
\bauthor{\bsnm{Liu}, \binits{N.}},
\bauthor{\bsnm{Hu}, \binits{X.}}:
\batitle{Techniques for interpretable machine learning}.
\bjtitle{Communications of the ACM}
\bvolume{63}(\bissue{1}),
\bfpage{68}--\blpage{77}
(\byear{2019})
\end{barticle}
\endbibitem

%%% 16
\bibitem{roscher2020explainable}
\begin{barticle}
\bauthor{\bsnm{Roscher}, \binits{R.}},
\bauthor{\bsnm{Bohn}, \binits{B.}},
\bauthor{\bsnm{Duarte}, \binits{M.F.}},
\bauthor{\bsnm{Garcke}, \binits{J.}}:
\batitle{Explainable machine learning for scientific insights and discoveries}.
\bjtitle{Ieee Access}
\bvolume{8},
\bfpage{42200}--\blpage{42216}
(\byear{2020})
\end{barticle}
\endbibitem

%%% 17
\bibitem{mott2019towards}
\begin{barticle}
\bauthor{\bsnm{Mott}, \binits{A.}},
\bauthor{\bsnm{Zoran}, \binits{D.}},
\bauthor{\bsnm{Chrzanowski}, \binits{M.}},
\bauthor{\bsnm{Wierstra}, \binits{D.}},
\bauthor{\bsnm{Jimenez~Rezende}, \binits{D.}}:
\batitle{Towards interpretable reinforcement learning using attention augmented
  agents}.
\bjtitle{Advances in Neural Information Processing Systems}
\bvolume{32},
\bfpage{12350}--\blpage{12359}
(\byear{2019})
\end{barticle}
\endbibitem

%%% 18
\bibitem{liu2018toward}
\begin{bchapter}
\bauthor{\bsnm{Liu}, \binits{G.}},
\bauthor{\bsnm{Schulte}, \binits{O.}},
\bauthor{\bsnm{Zhu}, \binits{W.}},
\bauthor{\bsnm{Li}, \binits{Q.}}:
\bctitle{Toward interpretable deep reinforcement learning with linear model
  u-trees}.
In: \bbtitle{Joint European Conference on Machine Learning and Knowledge
  Discovery in Databases},
pp. \bfpage{414}--\blpage{429}
(\byear{2018}).
\bcomment{Springer}
\end{bchapter}
\endbibitem

%%% 19
\bibitem{stiglic2020interpretability}
\begin{barticle}
\bauthor{\bsnm{Stiglic}, \binits{G.}},
\bauthor{\bsnm{Kocbek}, \binits{P.}},
\bauthor{\bsnm{Fijacko}, \binits{N.}},
\bauthor{\bsnm{Zitnik}, \binits{M.}},
\bauthor{\bsnm{Verbert}, \binits{K.}},
\bauthor{\bsnm{Cilar}, \binits{L.}}:
\batitle{Interpretability of machine learning-based prediction models in
  healthcare}.
\bjtitle{Wiley Interdisciplinary Reviews: Data Mining and Knowledge Discovery}
\bvolume{10}(\bissue{5}),
\bfpage{1379}
(\byear{2020})
\end{barticle}
\endbibitem

%%% 20
\bibitem{rutherford2004effect}
\begin{barticle}
\bauthor{\bsnm{Rutherford}, \binits{M.D.}}:
\batitle{The effect of social role on theory of mind reasoning}.
\bjtitle{British Journal of Psychology}
\bvolume{95}(\bissue{1}),
\bfpage{91}--\blpage{103}
(\byear{2004})
\end{barticle}
\endbibitem

%%% 21
\bibitem{perez2020adopting}
\begin{barticle}
\bauthor{\bsnm{Perez-Osorio}, \binits{J.}},
\bauthor{\bsnm{Wykowska}, \binits{A.}}:
\batitle{Adopting the intentional stance toward natural and artificial agents}.
\bjtitle{Philosophical Psychology}
\bvolume{33}(\bissue{3}),
\bfpage{369}--\blpage{395}
(\byear{2020})
\end{barticle}
\endbibitem

%%% 22
\bibitem{firestone2020performance}
\begin{barticle}
\bauthor{\bsnm{Firestone}, \binits{C.}}:
\batitle{Performance vs. competence in human--machine comparisons}.
\bjtitle{Proceedings of the National Academy of Sciences}
\bvolume{117}(\bissue{43}),
\bfpage{26562}--\blpage{26571}
(\byear{2020})
\end{barticle}
\endbibitem

%%% 23
\bibitem{chakraborti2019explicability}
\begin{bchapter}
\bauthor{\bsnm{Chakraborti}, \binits{T.}},
\bauthor{\bsnm{Kulkarni}, \binits{A.}},
\bauthor{\bsnm{Sreedharan}, \binits{S.}},
\bauthor{\bsnm{Smith}, \binits{D.E.}},
\bauthor{\bsnm{Kambhampati}, \binits{S.}}:
\bctitle{Explicability? legibility? predictability? transparency? privacy?
  security? the emerging landscape of interpretable agent behavior}.
In: \bbtitle{Proceedings of the International Conference on Automated Planning
  and Scheduling},
vol. \bseriesno{29},
pp. \bfpage{86}--\blpage{96}
(\byear{2019})
\end{bchapter}
\endbibitem

%%% 24
\bibitem{macnally2018action}
\begin{bchapter}
\bauthor{\bsnm{MacNally}, \binits{A.M.}},
\bauthor{\bsnm{Lipovetzky}, \binits{N.}},
\bauthor{\bsnm{Ramirez}, \binits{M.}},
\bauthor{\bsnm{Pearce}, \binits{A.R.}}:
\bctitle{Action selection for transparent planning}.
In: \bbtitle{AAMAS},
pp. \bfpage{1327}--\blpage{1335}
(\byear{2018})
\end{bchapter}
\endbibitem

%%% 25
\bibitem{chakraborti2017plan}
\begin{bchapter}
\bauthor{\bsnm{Chakraborti}, \binits{T.}},
\bauthor{\bsnm{Sreedharan}, \binits{S.}},
\bauthor{\bsnm{Zhang}, \binits{Y.}},
\bauthor{\bsnm{Kambhampati}, \binits{S.}}:
\bctitle{Plan explanations as model reconciliation: Moving beyond explanation
  as soliloquy}.
In: \bbtitle{26th International Joint Conference on Artificial Intelligence,
  IJCAI 2017},
pp. \bfpage{156}--\blpage{163}
(\byear{2017}).
\bcomment{International Joint Conferences on Artificial Intelligence}
\end{bchapter}
\endbibitem

%%% 26
\bibitem{bied2020integrating}
\begin{bchapter}
\bauthor{\bsnm{Bied}, \binits{M.}},
\bauthor{\bsnm{Chetouani}, \binits{M.}}:
\bctitle{Integrating an observer in interactive reinforcement learning to learn
  legible trajectories}.
In: \bbtitle{2020 29th IEEE International Conference on Robot and Human
  Interactive Communication (RO-MAN)},
pp. \bfpage{760}--\blpage{767}
(\byear{2020}).
\bcomment{IEEE}
\end{bchapter}
\endbibitem

%%% 27
\bibitem{zhao2020actor}
\begin{bchapter}
\bauthor{\bsnm{Zhao}, \binits{X.}},
\bauthor{\bsnm{Fan}, \binits{T.}},
\bauthor{\bsnm{Wang}, \binits{D.}},
\bauthor{\bsnm{Hu}, \binits{Z.}},
\bauthor{\bsnm{Han}, \binits{T.}},
\bauthor{\bsnm{Pan}, \binits{J.}}:
\bctitle{An actor-critic approach for legible robot motion planner}.
In: \bbtitle{2020 IEEE International Conference on Robotics and Automation
  (ICRA)},
pp. \bfpage{5949}--\blpage{5955}
(\byear{2020}).
\bcomment{IEEE}
\end{bchapter}
\endbibitem

%%% 28
\bibitem{kostrikov2021offline}
\begin{bchapter}
\bauthor{\bsnm{Kostrikov}, \binits{I.}},
\bauthor{\bsnm{Fergus}, \binits{R.}},
\bauthor{\bsnm{Tompson}, \binits{J.}},
\bauthor{\bsnm{Nachum}, \binits{O.}}:
\bctitle{Offline reinforcement learning with fisher divergence critic
  regularization}.
In: \bbtitle{International Conference on Machine Learning},
pp. \bfpage{5774}--\blpage{5783}
(\byear{2021}).
\bcomment{PMLR}
\end{bchapter}
\endbibitem

%%% 29
\bibitem{wu2019behavior}
\begin{botherref}
\oauthor{\bsnm{Wu}, \binits{Y.}},
\oauthor{\bsnm{Tucker}, \binits{G.}},
\oauthor{\bsnm{Nachum}, \binits{O.}}:
Behavior regularized offline reinforcement learning.
arXiv preprint arXiv:1911.11361
(2019)
\end{botherref}
\endbibitem

%%% 30
\bibitem{mysore2021regularizing}
\begin{bchapter}
\bauthor{\bsnm{Mysore}, \binits{S.}},
\bauthor{\bsnm{Mabsout}, \binits{B.}},
\bauthor{\bsnm{Mancuso}, \binits{R.}},
\bauthor{\bsnm{Saenko}, \binits{K.}}:
\bctitle{Regularizing action policies for smooth control with reinforcement
  learning}.
In: \bbtitle{2021 IEEE International Conference on Robotics and Automation
  (ICRA)},
pp. \bfpage{1810}--\blpage{1816}
(\byear{2021}).
\bcomment{IEEE}
\end{bchapter}
\endbibitem

%%% 31
\bibitem{szepesvari2010algorithms}
\begin{barticle}
\bauthor{\bsnm{Szepesv{\'a}ri}, \binits{C.}}:
\batitle{Algorithms for reinforcement learning}.
\bjtitle{Synthesis lectures on artificial intelligence and machine learning}
\bvolume{4}(\bissue{1}),
\bfpage{1}--\blpage{103}
(\byear{2010})
\end{barticle}
\endbibitem

%%% 32
\bibitem{brockman2016openai}
\begin{botherref}
\oauthor{\bsnm{Brockman}, \binits{G.}},
\oauthor{\bsnm{Cheung}, \binits{V.}},
\oauthor{\bsnm{Pettersson}, \binits{L.}},
\oauthor{\bsnm{Schneider}, \binits{J.}},
\oauthor{\bsnm{Schulman}, \binits{J.}},
\oauthor{\bsnm{Tang}, \binits{J.}},
\oauthor{\bsnm{Zaremba}, \binits{W.}}:
Openai gym.
arXiv preprint arXiv:1606.01540
(2016)
\end{botherref}
\endbibitem

\end{thebibliography}

\end{document}